\documentclass[letterpaper]{article} 
\usepackage{aaai24}  
\usepackage{times}  
\usepackage{helvet}  
\usepackage{courier}  
\usepackage[hyphens]{url}  
\usepackage{graphicx} 
\usepackage{natbib}  
\usepackage{caption} 
\usepackage{algorithm}
\usepackage{algorithmic}
\usepackage{amsmath}
\usepackage{amssymb}
\usepackage{multirow}
\usepackage{booktabs}
\usepackage{amssymb}
\usepackage{graphicx} 
\usepackage{subcaption} 
\usepackage{newfloat}
\usepackage{listings}
\DeclareCaptionStyle{ruled}{labelfont=normalfont,labelsep=colon,strut=off} 
\floatstyle{ruled}
\newfloat{listing}{tb}{lst}{}
\floatname{listing}{Listing}
\usepackage{color}

\title{Exploiting Facial Relationships and Feature Aggregation \\for Multi-Face Forgery Detection}
\author{
    Chenhao Lin \textsuperscript{\rm 1} ,Fangbin Yi\textsuperscript{\rm 1} ,Hang Wang\textsuperscript{\rm 1},Qian Li\textsuperscript{\rm 1},Deng Jingyi\textsuperscript{\rm 1},Chao Shen\textsuperscript{\rm 1}
}
\affiliations{
    \textsuperscript{\rm 1}Faculty of Electronic and Information Engineering, Xi’an Jiaotong University\\

}

\usepackage{bibentry}

\begin{document}

\maketitle

\begin{abstract}
 Face forgery techniques have emerged as a forefront concern, and numerous detection approaches have been proposed to address this challenge. However, existing methods predominantly concentrate on single-face manipulation detection, leaving the more intricate and realistic realm of multi-face forgeries relatively unexplored. This paper proposes a novel framework explicitly tailored for multi-face forgery detection,
 filling a critical gap in the current research. The framework mainly involves two modules: (i) a facial relationships learning module, which generates distinguishable local features for each face within images,
 (ii) a global feature aggregation module that leverages the mutual constraints between global and local information to enhance forgery detection accuracy. 
 Our experimental results on two publicly available multi-face forgery datasets demonstrate that the proposed approach achieves state-of-the-art performance in multi-face forgery detection scenarios. 

\end{abstract}

\section{Introduction}
In recent years, the rapid development of deepfake technology has achieved impressive results in generating synthetic multimedia contents \cite{Deepfakes,thies2019deferred,thies2016face2face,Deepfakes1}. However, these technologies have also been maliciously abused, resulting in the creation of fake news \cite{huh2018fighting,agarwal2019protecting} or social media fraud. The widespread use of deepfakes poses evolving threats to the public across societal domains, law enforcement, and even national security. To address these threats, the academic community is actively engaged in researching deepfake detection technology. 

As deepfake technology advances and the demand for their applications in complex real-world scenarios increases, there is a growing trend in research from single-face forgery to multi-face scenarios \cite{FFIW,le2021openforensics}. 
These more realistic scenarios pose new challenges to forgery detection research since the number of real and manipulated faces, as well as the various forgery techniques used, are typically unknown to defenders.

Although there are numerous studies on face forgery detection, most existing methods are designed for single-face forgery detection \cite{afchar2018mesonet,dang2020detection,guera2018deepfake,zhao2021multi,jung2020deepvision,nguyen2019multi,kumar2020detecting}. These methods can be adapted to multi-face forgery detection by detecting each face in the images first, and then identifying whether there are artifacts present. However, directly applying these methods in multi-face scenarios may not be effective because they only focus on individual faces and disregard underlying information that could help detect forgery. 

More recently, a few attempts, such as S-MIL \cite{S-MIL} and FFIW \cite{FFIW} have been made to address forgery detection in multi-face scenarios. These studies generally adopt a multiple-instance learning framework and train models using video-level labels for multi-face forgery detection in the video level. \textcolor{black}{However, they do not fully exploit the relationships between faces or global information in the images, and their detection accuracy still falls short of the requirements for large-scale applications}.

\begin{figure}[t!]
  \centering
  \includegraphics[width=\linewidth]{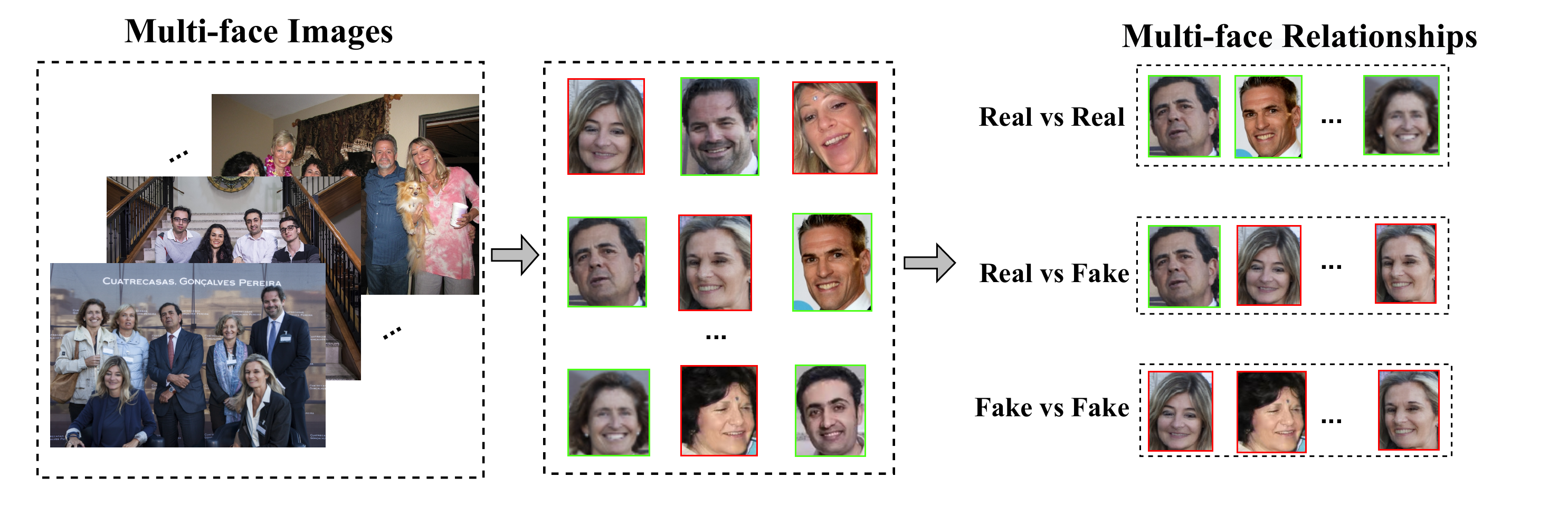}
  \caption{The faces in different multi-face images are extracted and their relationships (i.e., real vs real, real vs fake, and fake vs fake) are expected to be learned.}
\end{figure}

To address the \textcolor{black}{abovementioned} limitations and bridge the existing research gap, this paper introduces a novel framework, named FILTER, for multi-face forgery detection that leverages the proposed \textbf{F}acial relat\textbf{I}onships \textbf{L}earning and fea\textbf{T}ur\textbf{E} agg\textbf{R}egation approach. 
Figure 1 illustrates the core motivation and idea of our approach. We attempt to build connections between faces and learn the facial relationships embedded in multiple faces in the images for multi-face forgery detection. 


Specifically, we design two novel modules for multi-face forgery detection.
Firstly, we propose a multi-face relationships learning module to capture the underlying relationships between different faces within each multi-face image, which has been overlooked in previous studies.
Secondly, we introduce a global feature aggregation module to integrate the local facial features with global information of each multi-face image. This integration facilitates mutual constraints and guidance between local and global features, enhancing the detection of multi-face forgery. 
The experimental results on two publicly available multi-face forgery datasets Openforensics \cite{le2021openforensics} and FFIW10K \cite{FFIW}, validate the effectiveness of our FILTER on both face-level and image-level forgery detection in multi-face scenarios.
Overall, our contributions can be summarized into three aspects:
\begin{itemize}
    \item We propose a novel forgery detection framework for multi-face scenarios, which, to the best of our knowledge, is the first attempt leveraging facial relationships within images for image-level multi-face forgery detection field.
    \item 
    We design a multi-face relationship learning module to effectively capture the underlying relationships between different faces by constructing a self-similarity matrix. A global feature aggregation module is designed to achieve the image-level supervision and is integrated with local facial feature for enhanced detection performance. 


    \item Our proposed method exhibits promising results on two multi-face forgery datasets. In comparison to both single-face and multi-face detection methods, our method achieves a significant improvement in detecting multi-face forgery.
\end{itemize}

\section{Related Work}
\subsection{Multi-Face Deepfake}
The remarkable progress of deep learning has led to significant advancements in deepfake technology, with a particular focus on forgery faces generation. Variational Autoencoders (VAEs) \cite{hou2017deep,sinha2019variational} and Generative Adversarial Networks (GANs) \cite{arjovsky2017wasserstein,zhang2019self,mao2017least,karras2019style} have emerged as prominent synthesis techniques, contributing to the creation of high-quality and lifelike facial images. Currently, most research on face forgery focuses on single face manipulation and the primary objective is to achieve more realistic forgery effects that are challenging for both human eyes and detectors to differentiate.

Recently, there has been an increasing interest in the forgery of multiple faces, which is more applicable to real-world scenarios. Multiple face forgery involves manipulating some or all of the faces in an image or video containing multiple faces, posing a greater challenge for forgery detection. Existing research on multiple face forgery includes OpenForensics \cite{le2021openforensics}, which gathers raw images of multiple faces, extracts the latent vectors of each face's identity, iteratively modifies the latent vectors with random values, and feeds them into a GAN model to generate new faces. The feasible operation regions are then extracted and mixed with the original faces to form new identities until a simple classifier can be deceived. Another study on multiple face forgery is FFIW \cite{FFIW}, which develops a Domain-Adversarial Quality Assessment Network (Q-Net) to evaluate the quality of faces generated by various generative models (such as StyleGAN \cite{karras2019style}, StyleGAN2 \cite{karras2020analyzing}, and PGGAN \cite{karras2017progressive}) at different iteration stages. The iteration stages are used as pseudo-labels to train the quality evaluation model. Finally, Q-Net is combined with three existing face swapping methods(DeepFaceLab \cite{perov2020deepfacelab}, FS-GAN \cite{nirkin2019fsgan}, and FaceSwap \cite{Deepfakes1}) to automatically construct the $FFIW_{10K}$ multi-face forgery dataset, ensuring the quality of the forged faces. These techniques offer low-cost and efficient ways to synthesize a large number of high-quality multi-face forgery images, thereby presenting new challenges to deepfake detection field.

\begin{figure*}[t]
  \includegraphics[width=\textwidth]{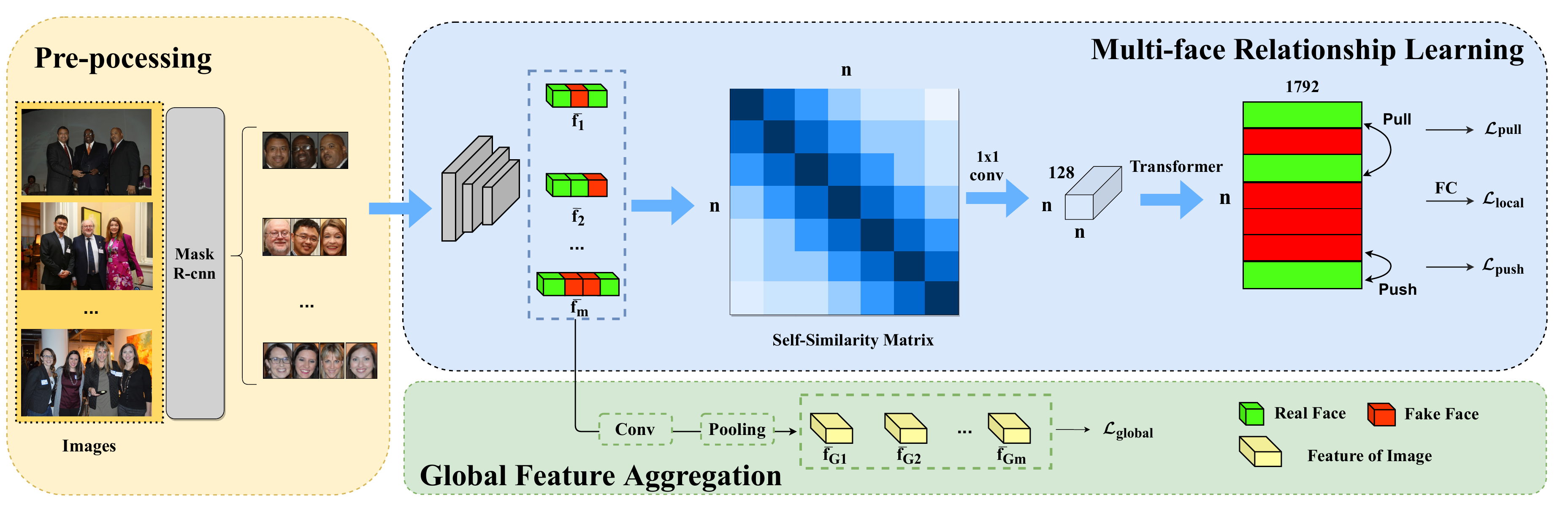}
  \caption{FILTER Framework Overview. Features are extracted from all individual faces in multi-face images, followed by the Multi-face Relationship Learning module, which learns the interrelationships among the faces. Simultaneously, facial features in each multi-face image are globally aggregated to obtain overall characteristics, assisting in the detection of forgery faces. }
  \label{Fig2}
\end{figure*}

\subsection{Deepfake Detection}
The rapid development of deepfake technology has prompted active research in related detection techniques. Currently, many methods focus on studying forgery detection from various perspectives. Some methods concentrate on specific features, such as analyzing eye blinking \cite{jung2020deepvision}, head posture \cite{yang2019exposing}, mouth movement \cite{haliassos2021lips}, and the consistency of corneal specular highlights \cite{hu2021exposing}. Other methods focus on the frequency domain \cite{li2021frequency,liu2021spatial}, capturing artifacts introduced by GANs in the spectrum and the frequency domain feature differences between real and fake faces. Nevertheless, the most optimal offen approaches rely on various methods that use convolutional neural networks, such as Xception and EfficientNet, to capture forgery clues \cite{rossler2019faceforensics++,nguyen2019multi,nirkin2021deepfake}. In contrast to the methods mentioned above, SBI \cite{SBI} method blends pseudo source and target images from single pristine images to generate self-blended images, encouraging the classifier to learn more general and robust representations to identify more challenging fake samples.
Recent M2TR \cite{wang2022m2tr} adopts a two-stream architecture comprising a multi-scale transformer and frequency filters to capture forgery patterns in both spatial and frequency domain. A cross modality fusion block is employed to amalgamate the complementary forgery features for detection. M2TR is demonstrated as one of the stat-of-the-art single-face forgery detection methods \cite{lin2022towards}.

Most existing forgery detection methods are designed for single-face scenarios, although these methods can be directly applied to multi-face forgery detection tasks by using a face detector, their performance is not satisfactory in this more challenging scenario. 
To address this challenge, some methods have actively explored forgery detection in multi-face scenarios. For example, S-MIL \cite{S-MIL} proposes a sharp multiple-instance learning strategy that directly maps instance embeddings to bag predictions and designs spatio-temporal encoding instances to model the inconsistencies within and between frames, thereby improving the detection performance. In FFIW \cite{FFIW}, the author designs a multi-face detection approach, which also focuses on forgery detection in multi-face videos. It summarizes the short-term, long-term, and global features of each person's face trajectory in the video to obtain robust and discriminative representations, adaptively aggregates them into video-level representations, and trains them using video-level labels. However, these detection methods for multi-face forgery essentially extract features independently for each face and do not use the relationships that exist in multi-face scenarios to assist detection. Moreover, these methods are mainly designed for video-level forgery detection and lack dedicated detection methods for image-level forgery. Our proposed method attempt to address these issues and significantly improve the detection performance. Please note that several recent studies \cite{zhang2023contrastive} have proposed enhancements for multi-face forgery detection. However, their research primarily concentrates on forgery localization, which does not align with the main focus of our problem.


\section{Methodology}
This paper aims to develop an efficient forgery detection framework capable of accurately identifying all manipulated faces in real-world multi-face scenarios. 
\textcolor{black}{A straightforward way is to individually detect each face in multi-face scenarios and apply existing single-face detection techniques directly. However, this strategy may not be effective in these scenarios, as it fails to consider the hidden correlation information among the faces in multi-face images.}
\textcolor{black}{To address this research gap, we present FILTER, a novel forgery detection framework that attempts to learn the relationships among different faces in images and aggregate local and global features for more accurate detection.} 

As illustrated in Figure \ref{Fig2}, FILTER mainly consists of three components: 
i) \textbf{Pre-processing} module, which detects all faces and generates visual features of each face in images, 
ii) \textbf{Multi-face Relationship Learning} module, which applies a self-similarity matrix followed by a Transformer encoder to learning discriminative features for each individual faces. In this module, a pull loss and a push loss are designed to improve the learned features,  
and iii) \textbf{Global Feature Aggregation} module, which uses the convolution with a pooling operation on features of all faces in an image to generate the global feature of each multi-face image.

\subsection{Pre-processing}
The first step is to detect each face in multi-face scenarios. Specifically, given a set of images, we assume each image $X$ consists of \textcolor{black}{$n$ faces}, i.e., 
\begin{math}
X =\{x_{1},x_{2},\ldots, x_n \}.
\end{math}
Then, we apply an object detector to detect faces and obtain the detection boxes for all faces in each image. We represent the $i-th$ face in the image as $x_i$, and its corresponding face label as $y_i$. Since face forgery detection is essentially a binary classification problem, its label information can be expressed as $y_i\in\{0,1\}$, where we set $y_i=0$ to represent a real face and $y_i=1$ to represent a fake face. The relationship between the global image-level label $y$ and the face-level label $y_i$ of each face can be represented as:
\begin{equation}
  y =\max\limits_{1 \leq t \leq n} y_{i}.
\end{equation}
After that, we extract features for all the detected faces as the next step. To achieve this, a pre-trained deep neural network serves as a feature extractor to obtain features $f_i$ for each detected face $x_i$. 


\subsection{Multi-face Relationship Learning}
To better capture the relationships between different faces in images, we propose to construct a self-similarity matrix to represent the distance between their visual features. In addition, we introduce metric learning to further enhance the discriminability of the captured feature learned from the self-similarity matrix.

\subsubsection{Self-similarity Matrix}
Specifically, for each feature $f_i$, we calculate the cosine distance between it and other facial features extracted from the detected faces in the whole dataset, to generate the self-similarity matrix $\mathbf{S} \in \mathbb{R}^{n \times n}$. 
After that, we utilize a $1 \times 1$ convolution module to enlarge the number of network channels, i.e., the number of self-similarity matrices, allowing the model to learn different correlation features from different similarity matrices. 
Then we apply a single-layer Transformer encoder to enhance the learning of facial correlation and generate the final local feature $F_i$ for each face. 
Finally, the local feature is fed into a fully-connected layer to predict the label $\hat{y}_{i}$ of the $i$-th face $x_{i}$ in the image $X$.

\subsubsection{Metric Learning}
Moreover, we propose to utilize metric learning to enhance the discriminability of extracted features. Given a real and forged face features $f_{Ri}$ and $f_{Fi}$ respectively in an image X, we design the pull loss and push loss, which are defined as:
\begin{equation}
    \mathcal L_{pull} = \sum_{i}^{N}(1-\cos(f_{Ri},F_{R}))+\sum_{j}^{M}(1-\cos(f_{Fj},F_{F}))
\end{equation}
\begin{equation}
    \mathcal L_{push} = e^{\sum_{i}^{M}-(1-\cos(f_{Fi},F_{R}))}+e^{\sum_{j}^{N}-(1-\cos(f_{Rj},F_{F}))},
\end{equation}
where $F_{R} $ and $F_{F}$ represent the average standard features of real and forged faces, respectively:
\begin{equation}
    F_{R}  = \frac{1}{N}\sum_{i=1}^{N}f_{Ri}
\end{equation}
\begin{equation}
    F_{F}  = \frac{1}{M}\sum_{j=1}^{M}f_{Fj}
\end{equation}
where $N$ and $M$ represent the number of real and forged faces, respectively, involved in constructing the similarity matrix.

\subsection{Global Feature Aggregation}
To utilize the global information of each multi-face image, we apply a convolution layer with pooling operation on all facial features in each multi-face image. The generated global feature of can be expressed as:
\begin{equation}
    \overline{f}_G = \mathtt{Conv(Pool} (\overline{f}_1, \overline{f}_2, ..., \overline{f}_j, ..., \overline{f}_m)),
\end{equation}
where $\mathtt{Conv}$ and $\mathtt{Pool}$ denote the convolution layer and pooling operation, respectively. $\overline{f}$ represents the features of all faces belong to a single multi-face image and $m$ is the number of images. Then, we feed the global feature $\overline{f}_G$ into a two-layer neural network to predict the image-level label $\hat{y}$, i.e., $\hat{y} = \mathtt {FC} (\overline{f}_G)$. In such way, the global feature will be combined with local facial feature effectively for more accurate detection. 

\subsection{Loss Function}
The total loss of our framework FILTER is made up of the global loss, local loss, pull loss and push loss.
The global loss $\mathcal L_{global}$ is defined as follows:
\begin{equation}
  \mathcal L_{global} =  -y \log \hat{y} - (1 - y) \log (1 - \hat{y}).
\end{equation}
where $y$ is the ground-truth label of the image. The local loss $\mathcal L_{local}$ is defined as
\begin{equation}
    \mathcal L_{local} = -y_{i} \log \hat{y}_{i} - (1-y_{i}) \log (1-\hat{y}_{i}).
\end{equation}
where $y_{i}$ is the ground-truth label of $x_{i}$.

Finally, the total loss is defined as follows:
\begin{equation}
    \mathcal L = \mathcal L_{global} + \lambda_1 \mathcal L_{local} + \lambda_2 \mathcal L_{pull} + \lambda_3 \mathcal L_{push}.
\end{equation}
where $\lambda_1$, $\lambda_2$ and $\lambda_3$ are weights balancing different losses.

\begin{table*}
 \centering
  \label{tab:freq}
  \begin{tabular}{lcccc}

    \toprule
    \multirow{2}{*}{Method}& \multicolumn{2}{c}{Test-Dev} & \multicolumn{2}{c}{Test-Challenge}\\
                        &AUC(\%) & ACC(\%) & AUC(\%) &  ACC(\%) \\
                        
    \midrule
    
    Xception \ & 98.26 &  94.67 &  81.19 &  73.64 \\
    EFNB4    \ & 99.29 &  95.55 &  81.71 & 72.81 \\
    SBI     \ &  98.76&  92.71  &  82.46& 74.69 \\
    Multi-attention  \ & 99.20 & \textbf{99.40}  & 86.02& 75.10 \\
   \textbf{FILTER} & 99.82 & 98.93 & 89.89 & 81.78 \\
    \hline
    \hline
     M2TR     \ & \underline{99.86} & 98.20  & \underline{96.36}  & \underline{84.35}\\
    \textbf{FILTER+M2TR} &  \textbf{99.88}& \underline{99.00}   &  \textbf{96.89}& \textbf{89.01}\\
    
    \bottomrule
    
 \end{tabular}
  \caption{Evaluation Results on the Test-Dev and Test-Challenge Subsets of the $Openforensics$ Dataset. Several baseline and SOTA methods are compared. Bold and underlined values correspond to the best and the second-best value, respectively. }
\end{table*}


\section{Experiments}
\subsection{Implementation Details}
\textbf{Preprocessing.} For the location of faces in multi-face images, any popular object detector can be used. In this work, we adopt a two-stage detection model, MaskRcnn \cite{he2017mask}, and fine-tun the model on the Openforensics \cite{le2021openforensics} dataset to make it better suited for face detection. As for the problem of assigning labels to the detected faces, we calculate the IOU value between the detected face bounding boxes and the ground truth boxes one by one, and assign the label information of the ground truth box with the maximum IOU value to the detected face. At the same time, we uniformly scale the detected faces to a size of 224x224 as the input of the detection model.

\noindent\textbf{Model Training.}  We employ the powerful EfficientNet-b4 (ENFB4) \cite{efficientnet} as backbone networks to extract features as the representation of each face in multi-face images.
All methods are trained using the Adam optimizer with a learning rate of 1e-4. For video-level datasets, we sample only 5 frames for training from each video. Additionally, during the training process, we only apply simple data augmentation by randomly flipping the images horizontally.

\begin{table}[t]
  \label{tab:freq}
  \begin{tabular}{c c c}
    \specialrule{1pt}{0pt}{0pt}
    \hline
    \multirow{2}{*}{Method }& \multicolumn{2}{c}{classiﬁcation} \\
        \cmidrule(lr){2-3}
            &AUC(\%) & ACC(\%)  \\
                        
    \specialrule{1pt}{0pt}{0pt}
    \hline 
    \multicolumn{3}{l}{video-level methods:using face-level labels} \\
    \hline
    \multicolumn{1}{c|}{I3D} &69.5  & 68.8  \\
    \hline
    \multicolumn{3}{l}{video-level methods:using video-level labels} \\
    \hline
    \multicolumn{1}{c|}{S-MIL } & 61.2 & 59.8  \\
    \multicolumn{1}{c|}{FFIW } &  70.9 & 69.4  \\
    \hline
    \multicolumn{3}{l}{face-level methods:using face-level labels} \\
    \hline
    \multicolumn{1}{c|} {Xception }  &  56.1 &  54.1 \\
    \multicolumn{1}{c|}{FWA }  &  63.1 &  60.2  \\
    \hline
    \multicolumn{3}{l}{image-level methods:using face-level and video-level labels} \\
    \hline
    \multicolumn{1}{c|}{\textbf{FILTER}} &  \textbf{87.0}& \textbf{82.5} \\
   \specialrule{1pt}{0pt}{0pt}
    \hline
 \end{tabular}
 \caption{The evaluation results for multi-face forgery detection on the $FFIW_{10K}$ dataset.}
\end{table}

\subsection{Experimental Setting}
\textbf{Dataset.} We utilize two recent public and high-quality multi-face forgery datasets, \textbf{$Openforensics$} \cite{le2021openforensics} and \textbf{$FFIW_{10K}$} \cite{FFIW}, for our primary training and testing.

\textbf{$Openforensics$} employs a low-cost automated face synthesis process to generate a large number of forged facial images with high resolution and high visual quality. It contains two test subsets (\textbf{Test-Dev Set and Test-Challenge Set}), where the Challenge subset has over 40,000 forgery images with an average of 2.9 faces per image, which is consistent with the multi-face forgery scenario. Moreover, the Challenge subset uses various enhancement techniques to increase the difficulty of the forgery classification task. 

\textbf{$FFIW_{10K}$}, on the other hand, is a video-level forgery dataset that uses three existing face forgery methods to generate diverse forged faces and employs a quality assessment network to retain high-quality forged faces. Since this dataset is at the video level, we sampled frames to apply our image-level method. The number of frames extracted from each video is determined by the number of faces in the video. The more faces or forged faces in the video, the more frames we extract. Furthermore, to enhance comparability, we make sure that the total number of frames extracted from \textbf{$FFIW_{10K}$} is close to the number of images in the challenge subset of \textbf{$Openforensics$}.

\noindent\textbf{Baseline on Openforensics dataset.}  We select four face-level forgery detection methods,  Xception \cite{xception}, ENFB4 \cite{efficientnet},  SBI \cite{SBI},and M2TR\cite{wang2022m2tr} as our baseline for comparison. 
Xception is a representative method for forgery detection, while EFNB4 is one of the state-of-the-art network structures for many computer vision tasks, including forgery detection. They exhibit remarkable performance across a majority of existing forgery datasets. 
Multi-attention \cite{zhao2021multi} and SBI are two different types and popular methods, which are widely used in deepfake detection evaluations.    
M2TR introduces a cross-modal fusion module that effectively combines information from RGB and frequency streams, which is a recently proposed state-of-the-art method for deepfake detection. 
It's important to note that we do not incorporate other detection methods, such as Face x-ray \cite{li2020face}, Patch-Xception \cite{chai2020makes}, 
and LRNet \cite{sun2021improving} in our experiments, since Multi-attention has demonstrated superior performance compared to these methods.


\begin{figure*}[htbp]
  \centering
  \begin{subfigure}[b]{0.33\textwidth}
    \includegraphics[width=\textwidth]{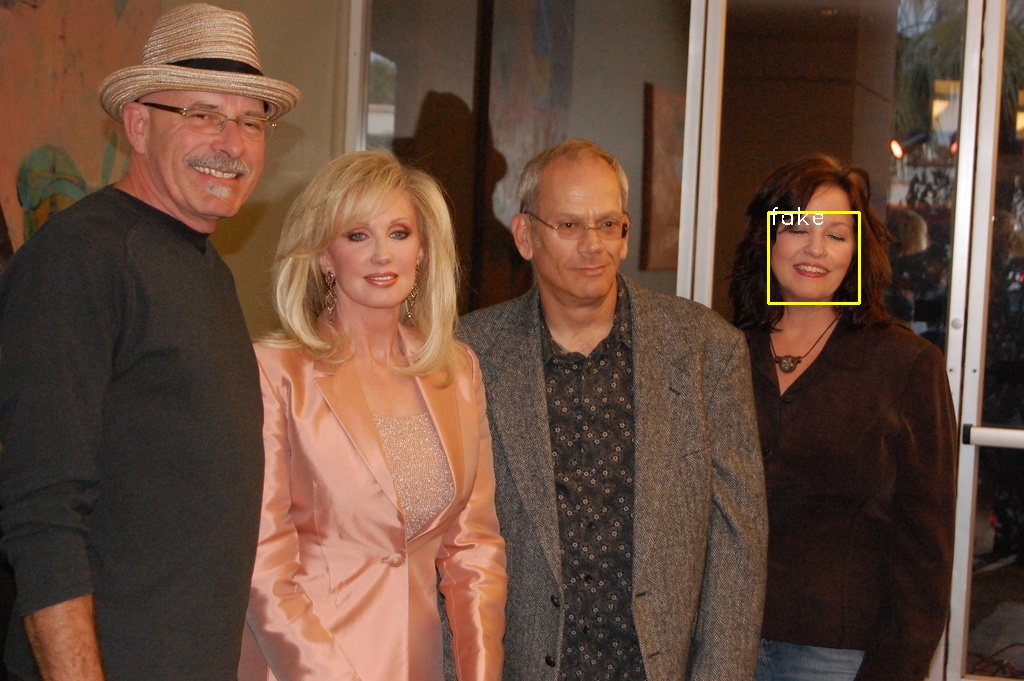}
    \label{fig:sub1}
  \end{subfigure}
  \begin{subfigure}[b]{0.33\textwidth}
    \includegraphics[width=\textwidth]{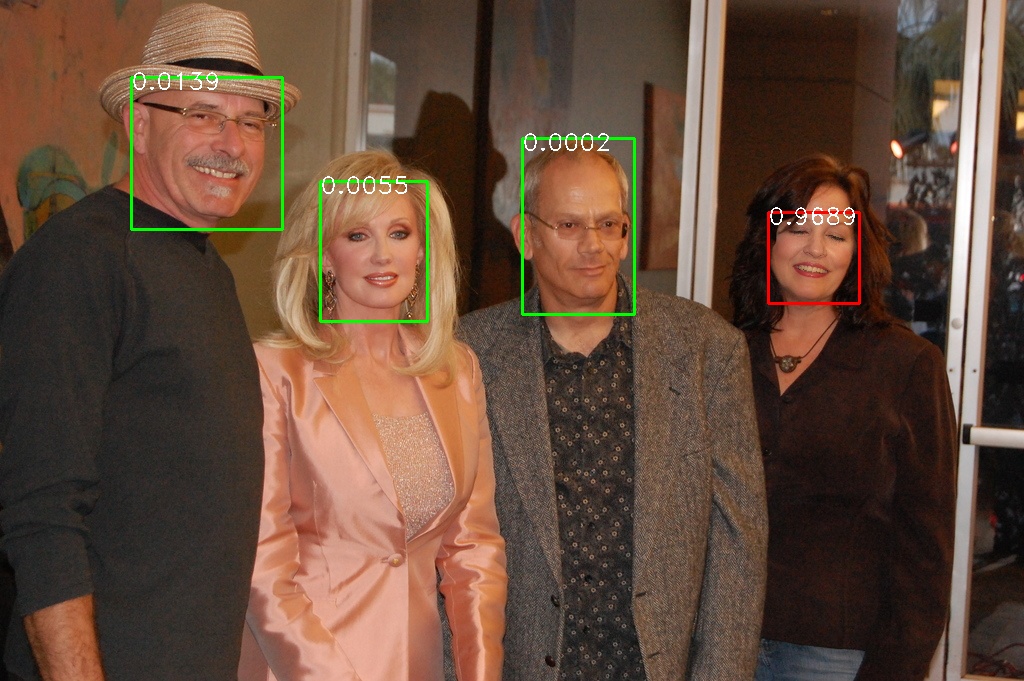}
    \label{fig:sub2}
  \end{subfigure}
  \begin{subfigure}[b]{0.33\textwidth}
    \includegraphics[width=\textwidth]{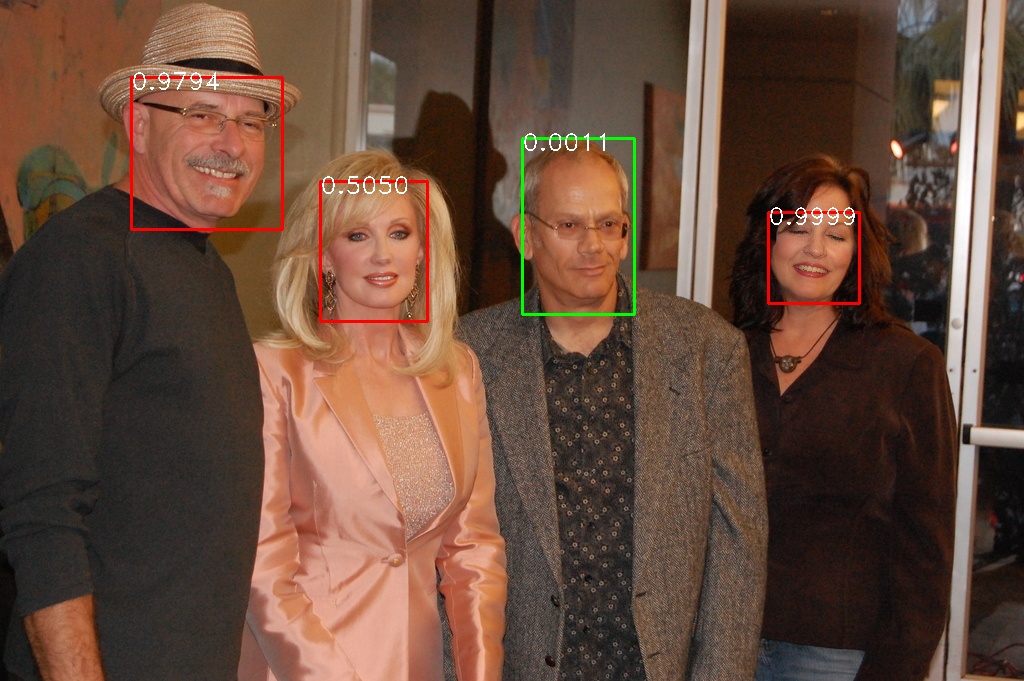}
    \label{fig:sub3}
  \end{subfigure}

  \begin{subfigure}[b]{0.33\textwidth}
    \includegraphics[width=\textwidth]{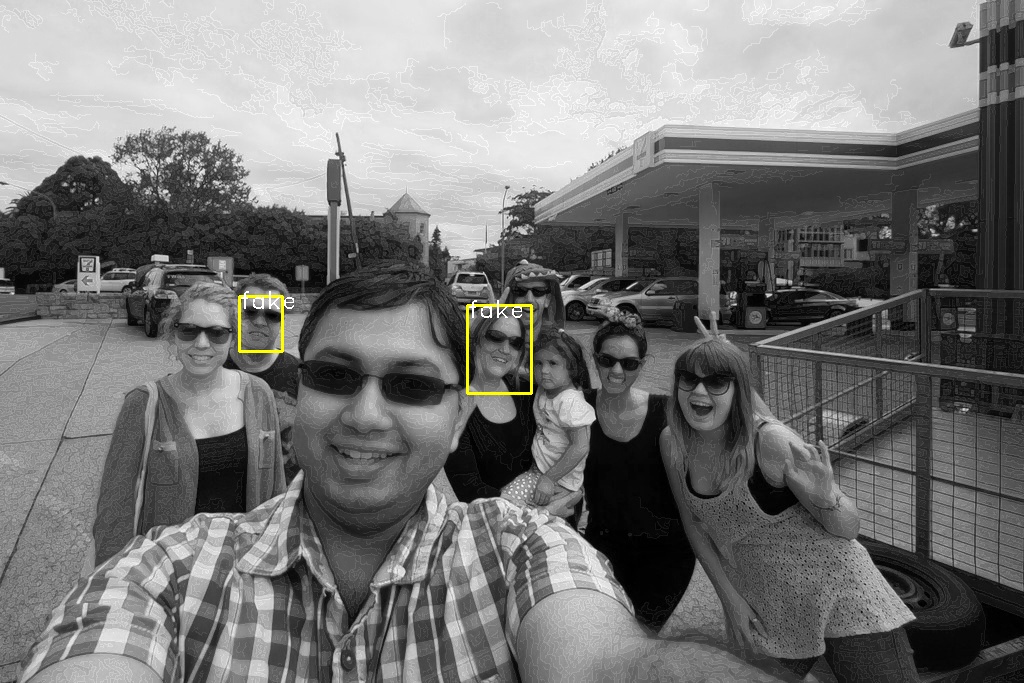}
    \label{fig:sub4}
    \caption{Original images}
  \end{subfigure}
  \begin{subfigure}[b]{0.33\textwidth}
    \includegraphics[width=\textwidth]{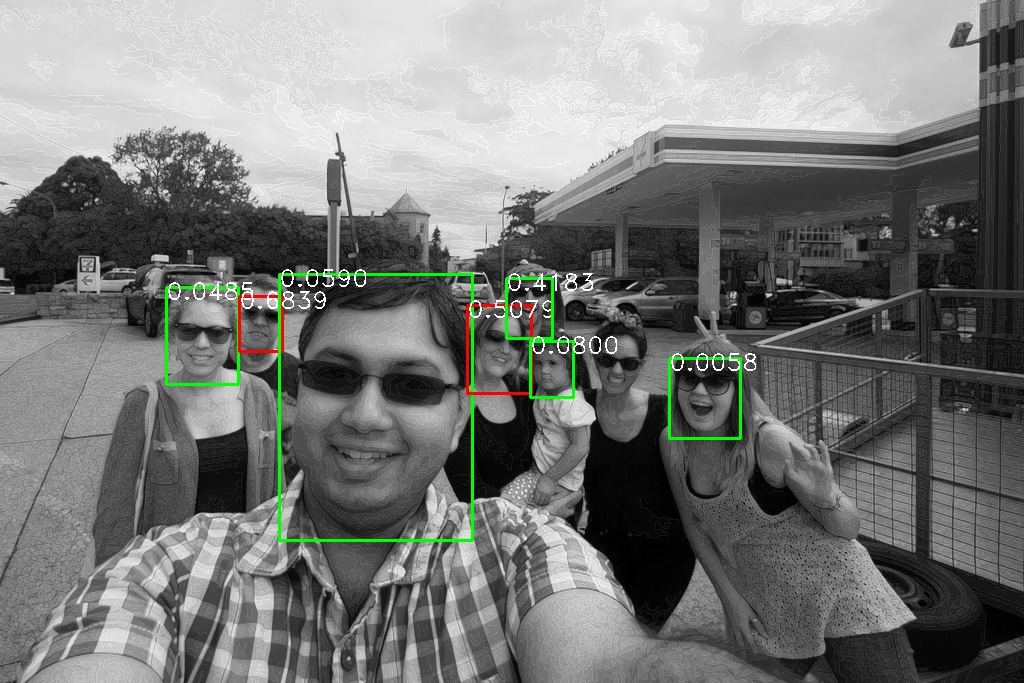}
    \label{fig:sub5}
    \caption{Detection results of our method}
  \end{subfigure}
  \begin{subfigure}[b]{0.33\textwidth}
    \includegraphics[width=\textwidth]{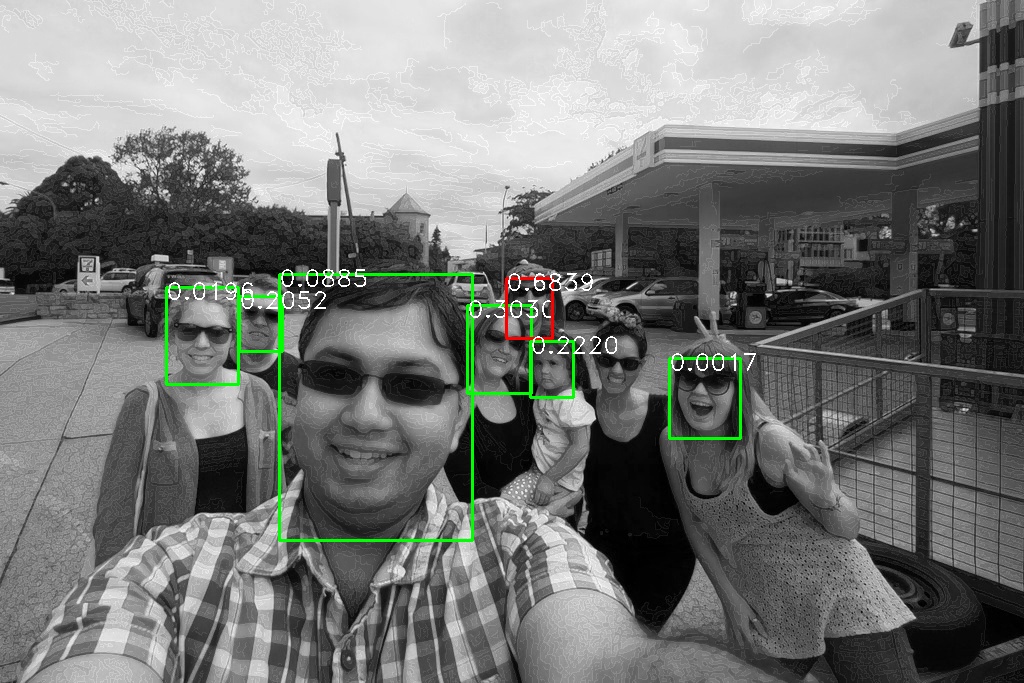}
    \label{fig:sub6}
    \caption{Detection results of Multi-attention}
  \end{subfigure}

  \caption{The comparative detection results on the $Openforensics$ Test-Dev set (first row) and Test-Challenge set (second row) between our method FILTER (b) and the Multi-attention method (c) are illustrated. The yellow box indicates that the face in the original image (a) is a forged face, the red box indicates that the face is detected as forged, and the green box indicates a real face. The number in the box represents the confidence score of the face being a forged face.}
  \label{}
\end{figure*}

\noindent\textbf{Baseline on FFIW10K dataset.}  On this dataset, we follow previous work \cite{FFIW} and select two face-level forgery detection methods, Xception \cite{xception} and FWA \cite{FWA}, as well as three video-level forgery detection methods, I3D \cite{I3D}, S-MIL \cite{S-MIL}, and FFIW, as our baselines. S-MIL and FFIW are designed for multiple-face scenarios, and FFIW achieves state-of-the-art performance on the current multiple-face forgery detection task.

\vspace{\baselineskip}
\noindent\textbf{Evaluation Protocols.} To conduct a comprehensive evaluation of our framework, we adopt two commonly used evaluation metrics in the field of deepfake detection: area under the receiver operating characteristic curve (AUC) and accuracy (ACC). For the evaluation on the $OpenForensics$ image-level dataset, we only compare our method with other face-level deepfake forgery methods, and both metrics are calculated at the face-level prediction score. For the evaluation on the $FFIW_{10K}$ video-level dataset, we compare our method with other face-level and video-level methods. The face-level methods process the video by sampling frames and perform detection on each face in each frame. Additionally, we follow the evaluation metrics in \cite{FFIW} to calculate the face-level AUC score and video-level accuracy (ACC) score for classification on the $FFIW_{10K}$ dataset.

\textcolor{black}{It is worth noting that the face-level detection methods we mentioned, such as Xception, refer to the methods that only perform forgery detection on each individual face in the image without considering the inter-relationship between the faces. On the other hand, our proposed method FILTER is an image-level detection method specifically designed to detect multiple faces in an image by considering other essential information between faces. Video-level detection methods refer to methods designed to detect video forgeries.}

\subsection{Experimental results on Openforensics}
We first conduct comparative experiments to evaluate the performance of our proposed method against baseline methods on the $Openforensics$ dataset.
All the methods use face-level labels as supervision, and the AUC and ACC scores are calculated at the face level. The comparative experimental results are illustrated in Table 1. 
The experimental results show that FILTER exhibits superior forgery detection performance on both test subsets. Compared to several baseline methods such as Xception, EFNB4, SBI, and Multi-attention, FILTER illustrates significant improvement on the Test-Challenge subset, with an average increase of 7.05\% in AUC score and 7.72\% in ACC score. This improvement can be attributed to the learned meaningful interrelationships among faces and aggregated global information by FILTER.

Meanwhile, to provide a more intuitive comparison between our method and the baseline Multi-attention method on multi-face images, we present a visualized detection performance comparison in Figure 3. The figure illustrates that our method can detect all forgery faces in the multi-face image without producing false positives.

In Table 1, M2TR illustrates superior performance on the Test-Challenge set due to its complex multi-stream transformer-based structure and combined features from both spatial and frequency domains. Nevertheless, FILTER can be seamlessly integrated into M2TR by concatenating the features extracted by M2TR with the features obtained from the similarity matrix of FILTER, achieving the best detection performance.


\begin{figure*}[t]
  \centering
  \begin{subfigure}[b]{0.33\textwidth}
    \includegraphics[width=\textwidth]{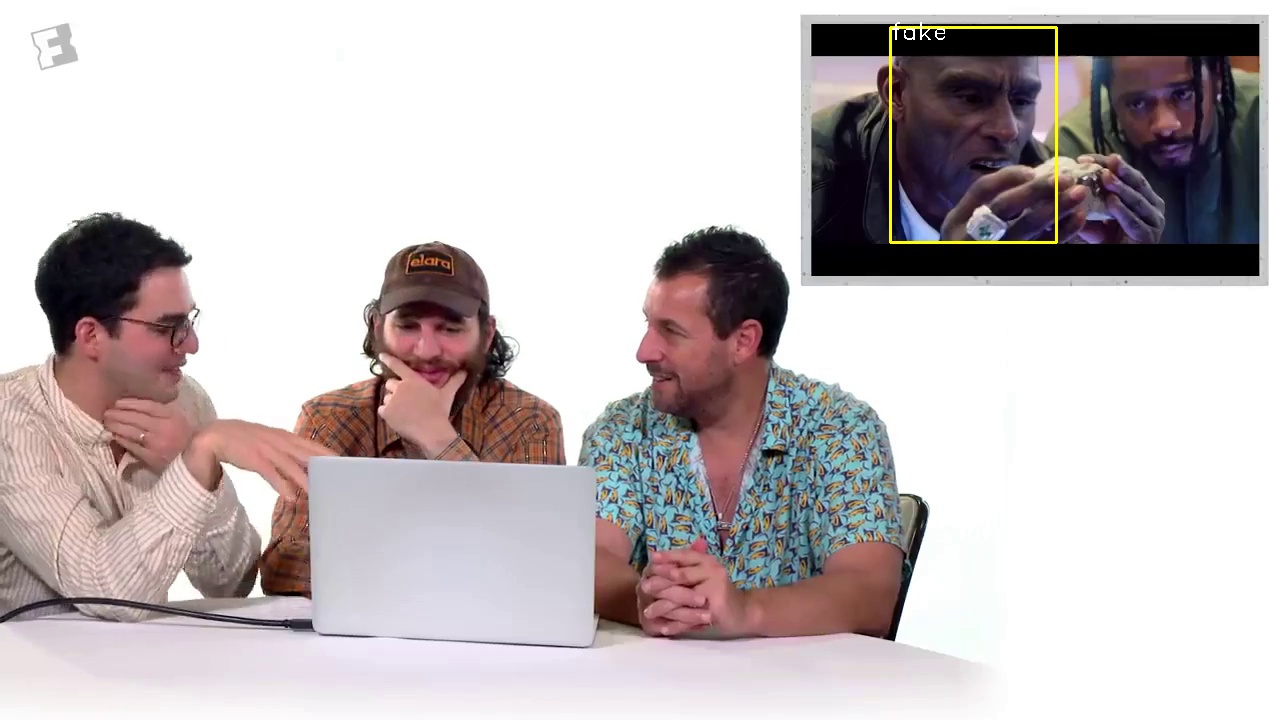}
    \label{fig:sub1}
  \end{subfigure}
  \begin{subfigure}[b]{0.33\textwidth}
    \includegraphics[width=\textwidth]{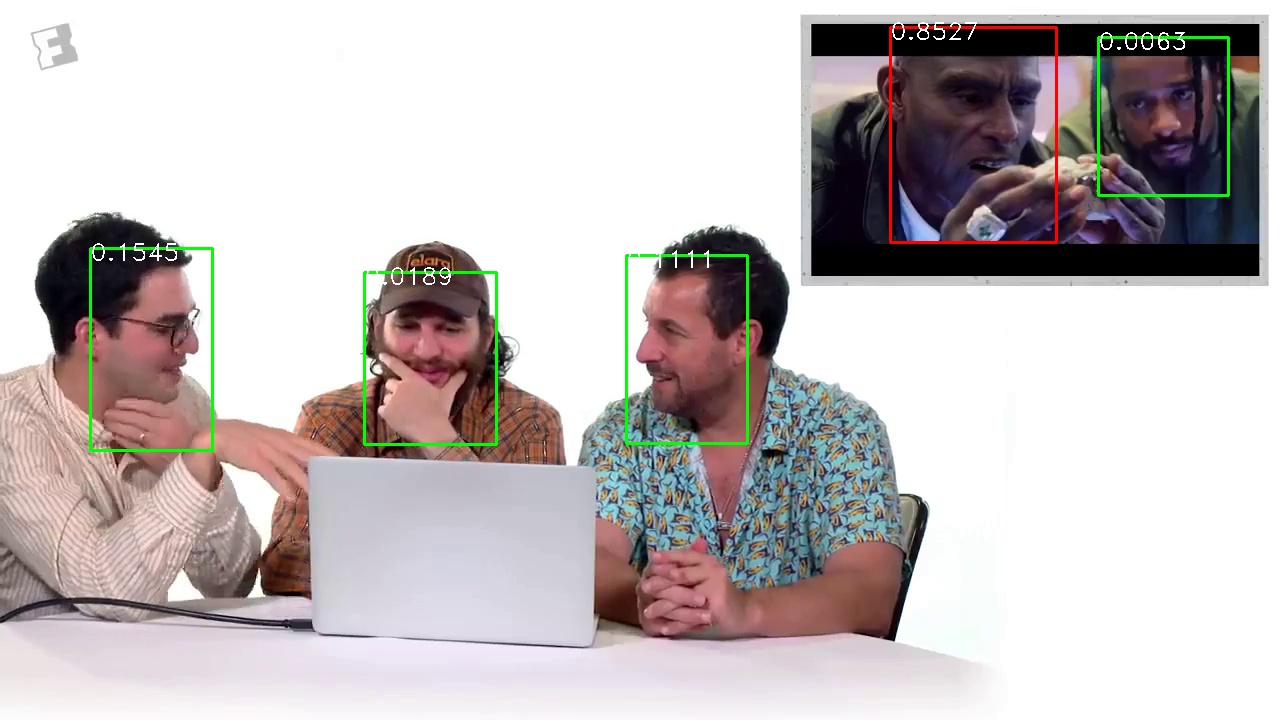}
    \label{fig:sub2}
  \end{subfigure}
  \begin{subfigure}[b]{0.33\textwidth}
    \includegraphics[width=\textwidth]{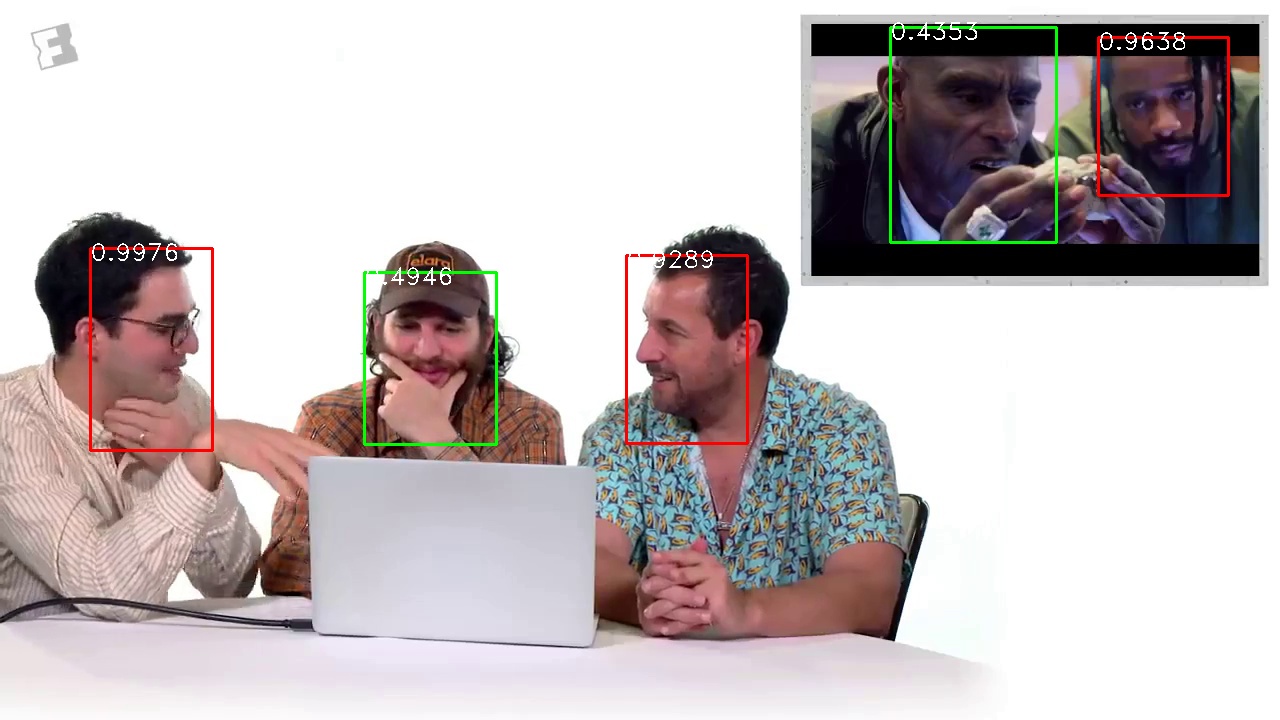}
    \label{fig:sub3}
  \end{subfigure}

  \begin{subfigure}[b]{0.33\textwidth}
    \includegraphics[width=\textwidth]{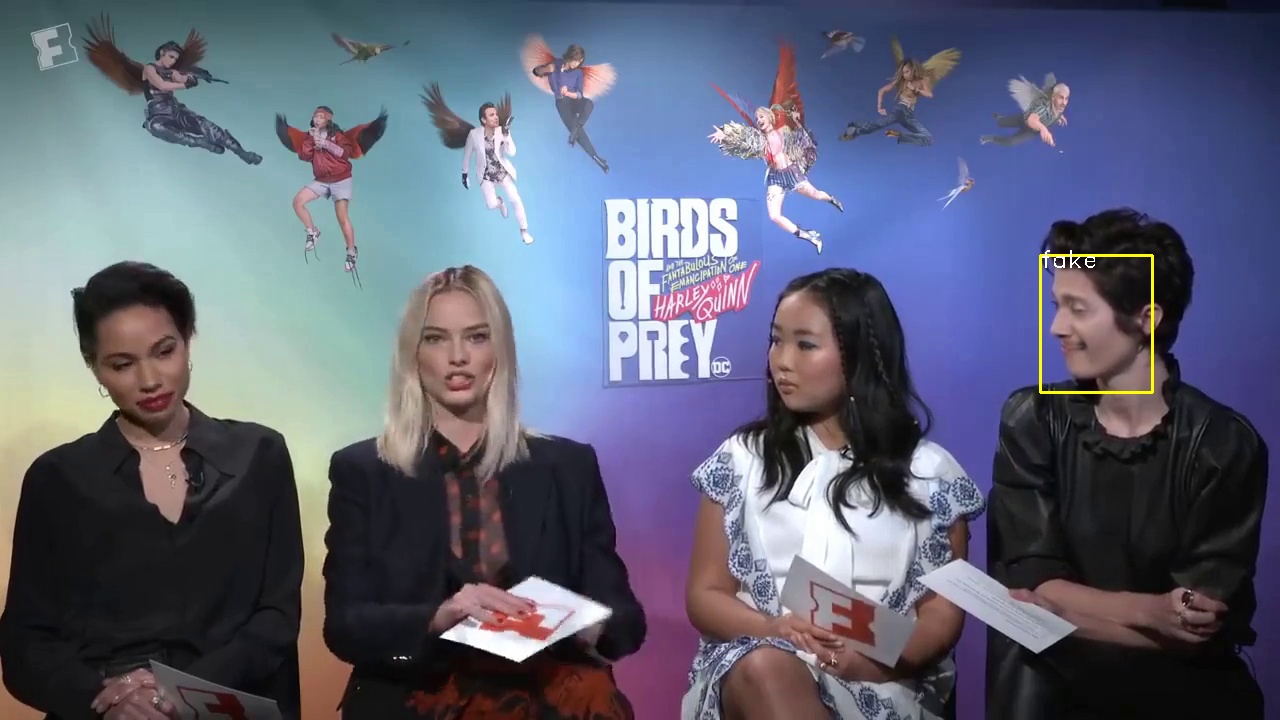}
    \label{fig:sub1}
    \caption{Original image}
  \end{subfigure}
  \begin{subfigure}[b]{0.33\textwidth}
    \includegraphics[width=\textwidth]{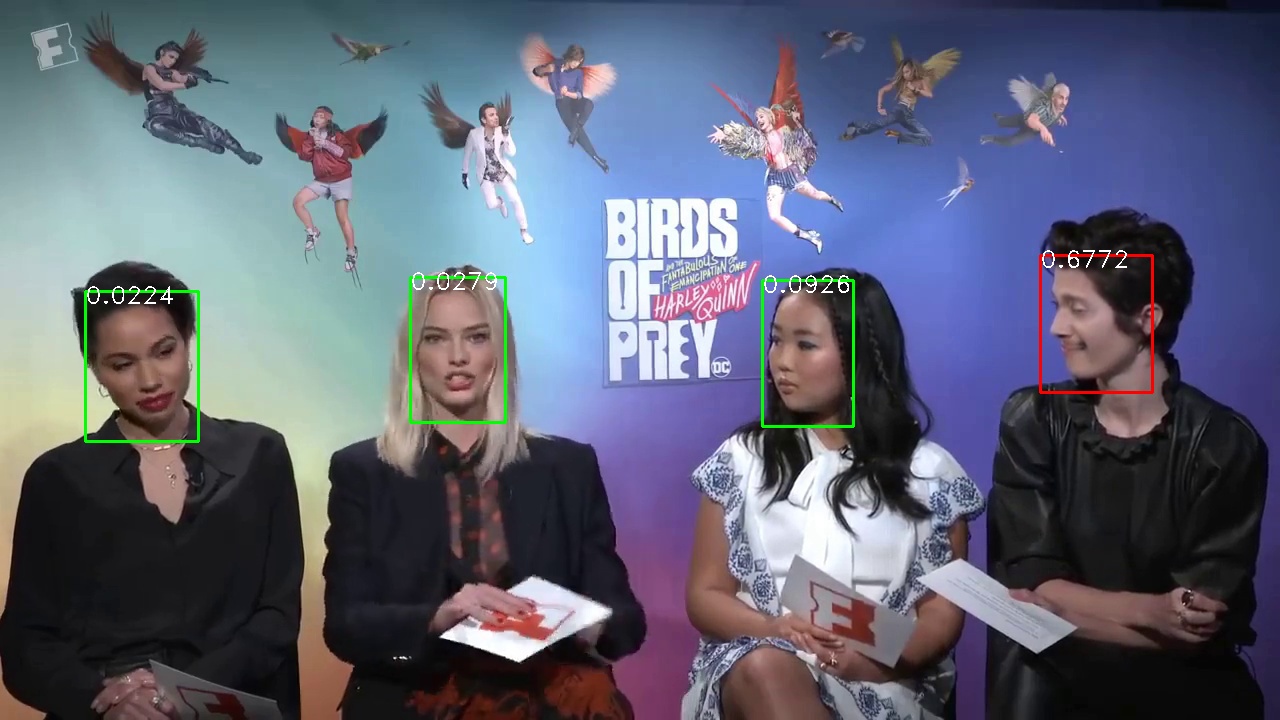}
    \label{fig:sub2}
    \caption{Detection result of our method}
  \end{subfigure}
  \begin{subfigure}[b]{0.33\textwidth}
    \includegraphics[width=\textwidth]{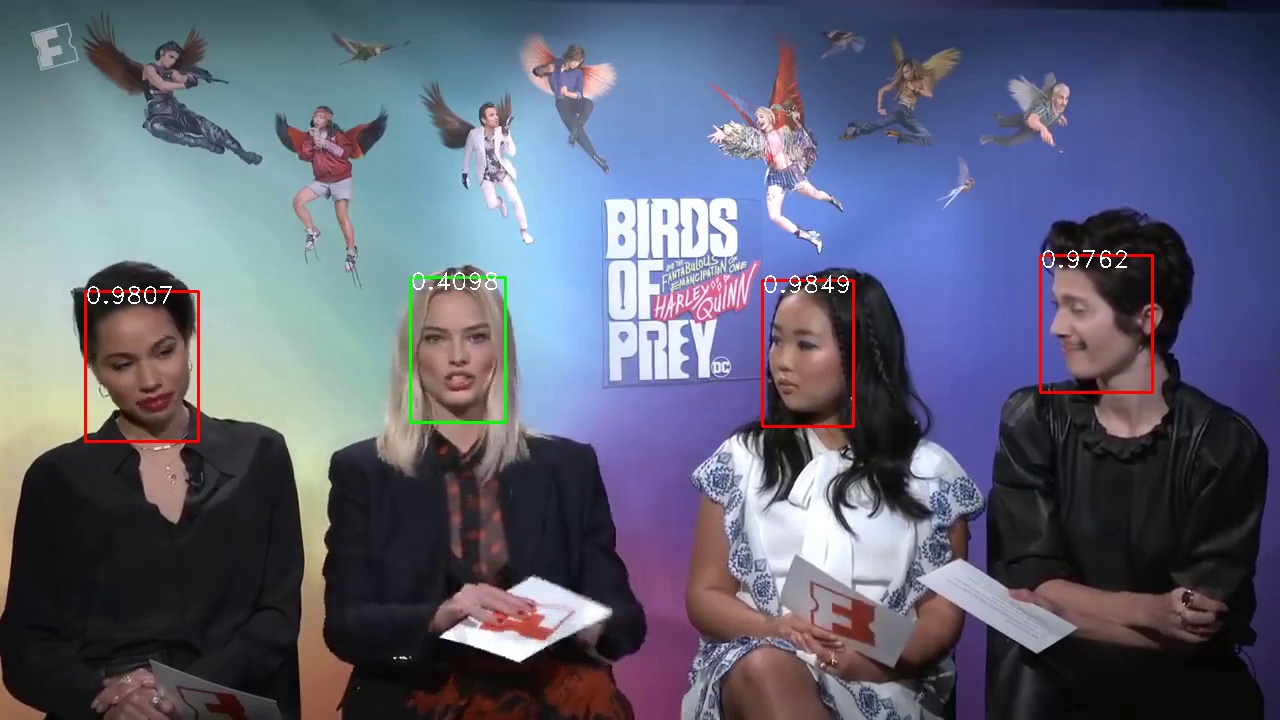}
    \label{fig:sub3}
    \caption{Detection result of the Xception}
  \end{subfigure}

  \caption{The comparative detection results on the $FFIW_{10K}$ test set between our method FILTER (b) and the Xception method (c) are illustrated. The yellow box indicates that the face in the original image (a) is a forged face, the red box indicates that the face is detected as a forged face, and the green box indicates a real face. The number in the box represents the confidence score of the face being a forged face.}
  \label{}
\end{figure*}

 \begin{table*}[t]
 \centering
  \label{tab:freq}
  \begin{tabular}{l|ccccc|cccc}
    \toprule
    \multirow{2}{*}{Aspect} & \multirow{2}*{ $ L_{global}$ } 
                          & \multirow{2}*{ $L_{local}$}
                          & \multirow{2}*{$L_{push}$}
                          & \multirow{2}*{$L_{pull}$}
                          & \multirow{2}*{SM}
                          & \multicolumn{2}{c}{Test-Dev(\%)} 
                          & \multicolumn{2}{c}{Test-Challenge(\%)}\\
    &  &  & & & &AUC &ACC &AUC &ACC\\

    \midrule
    No-global &  & \checkmark & \checkmark & \checkmark &\checkmark  
    &91.57 & 82.85 & 86.77 & 81.39  \\
    No-push     &\checkmark  &  \checkmark &  &\checkmark  & \checkmark  &99.81 & 97.63 & 88.08 & 81.43  \\
    No-pull  & \checkmark & \checkmark & \checkmark &  & \checkmark 
    &99.81 & \textbf{99.00} & 88.06 & 80.49   \\
    No-SM   & \checkmark &  \checkmark& \checkmark & \checkmark &  
    &98.57 & 93.97 & 84.03 & 76.01  \\
    \textbf{Full}   & \checkmark &  \checkmark& \checkmark & \checkmark & \checkmark  &\textbf{99.82} & 98.93 & \textbf{89.89} & \textbf{81.68}  \\
  \bottomrule
 \end{tabular}
 \caption{Ablation study on test set of Openforensics. "SM" represents the similarity matrix, Bold values correspond to the best value.}
\end{table*}

\begin{table}[t]
    \centering
    \begin{tabular}{c|cccc}
    \toprule
      \multirow{2}{*}{SM Size}  &  \multicolumn{2}{c}{Test-Dev(\%)} 
            & \multicolumn{2}{c}{Test-Challenge(\%)} \\
            & AUC  &ACC &AUC &ACC  \\
    \midrule
      10x10   & 99.38  &97.46 & 84.19  &  76.28  \\
      20x20   & 99.70  &97.55  & 85.27 &  77.84  \\
      30x30   &  99.78 & \textbf{99.03} & 87.03 &   79.31 \\
     \textbf{ 40x40 }  &  \textbf{99.82} & 98.93 &  \textbf{89.89}&   \textbf{81.68} \\
    \bottomrule
    \end{tabular}
    \caption{Ablation study on test set of Openforensics. The larger the dimensions of the similarity matrix, the higher the accuracy of detection.}
    \label{tab:my_label}
\end{table}

\vspace{\baselineskip}

\subsection{Experimental results on $FFIW_{10K}$}
To validate the effectiveness of our method on multiple datasets, we compare it with different face-level and video-level forgery detection methods on the $FFIW_{10K}$ dataset. Following the convention established in \cite{FFIW}, we evaluate our image-level method at the track-level by parsing each test video into a set of face trajectories and using the average score of all faces in each trajectory as its predicted score. Additionally, we report the video-level accuracy score (ACC) for classification. For the compared baselines, we follow the previous work \cite{FFIW} and directly use the original results reported in the paper. From the results shown in Table 2, we can observe that our method significantly outperforms all other compared methods. 
Unlike these methods, our approach utilizes both face-level and video-level labels as supervision. Moreover, our method exhibits more prominent performance in multi-face forgery detection tasks when compared to the FFIW \cite{FFIW} and S-MIL \cite{S-MIL} methods designed for the same task, achieving state-of-the-art results. Therefore, this indirectly confirms that utilizing the multi-face relationships and global information in this scenario to assist detection is indeed effective. Furthermore, in Figure 4, we illustrate the superior detection performance of our method compared to the face-level method Xception on the video-level forgery dataset.

\subsection{Ablation Studies}
To further evaluate the effectiveness of each part in our framework, we perform ablation studies on two test subsets from Openforensics, as shown in Table 3. 

Firstly, we conduct an ablation study on several key components constituting the framework. From the experimental results in Table 3, it is evident that the complete model achieves the best detection performance, with each individual component of the model learning information that contributes to forgery classification to some extent. Removing any of these components leads to a decrease in the detection performance. Notably, when the similarity matrix (SM) component is removed, the model experiences the most significant performance drop. This underscores the crucial role of the facial relationships learned by the similarity matrix in the FILTER framework.
Additionally, the global feature aggregation component also demonstrates a significant effect in assisting individual forgery classification.

Due to the significant impact of the similarity matrix on the model, we further investigate the influence of the similarity matrix size on the model's performance. From the experimental results in Table 4, it can be observed that as the size of the similarity matrix gradually increases, the detection performance of the model also improves correspondingly. We can achieve the best performance when the size of the similarity matrix equals to $40\times40$.

\section{Conclusion}
In this paper, we design the FILTER framework for multi-face forgery detection using a novel facial relationship learning and feature aggregation approach. 
A Multi-face Relationship Learning module is proposed to learn the associative information among faces, aiming to investigate the fundamental distinctions between real and forged faces. On the other hand, a Global Feature Aggregation module aggregates the features of all faces within the same multi-face image into a global feature, thereby assisting individual facial forgery classification from a global perspective. 
Our experimental results on two public datasets, including both face-level and image-level evaluations, demonstrate that FILTER can effectively detect multiple manipulated faces in a single image and significantly outperforms several baseline and state-of-the-art methods in multi-face forgery scenarios.

\bibliography{aaai24}

\end{document}